\documentclass[letterpaper, 10 pt, conference]{ieeeconf}  

\IEEEoverridecommandlockouts                              

\usepackage{graphics} 
\usepackage{epsfig} 
\usepackage{mathptmx} 
\usepackage{times} 
\usepackage{amsmath} 
\usepackage{amssymb}  
\usepackage{balance}

\title{\LARGE \bf Assessing the Contribution of Semantic Congruency to \\ Multisensory Integration and Conflict Resolution}

\author{Di Fu$^{1,2,3}$, Pablo Barros$^1$, German I. Parisi$^1$, Haiyan Wu$^{2,3,4}$, Sven Magg$^{1}$, Xun Liu$^{2,3}$, Stefan Wermter$^1$
\thanks{$^{1}$Knowledge Technology, Department of Informatics, Universit\"at Hamburg, Germany.
        {\tt\small \{surname\}@informatik.uni-hamburg.de}}%
\thanks{$^{2}$CAS Key Laboratory of Behavioral Science, Chinese Academy of Sciences (CAS), Beijing, China. {\tt\small \{fud,wuhy,liux\}@psych.ac.cn}}%
\thanks{$^{3}$Department of Psychology, University of CAS, Beijing, China.}%
\thanks{$^{4}$Division of the Humanities and Social Sciences, Caltech, CA, USA.}
\thanks{{\tt\small http://cml.knowledge-technology.info} }
}

\begin{document}

\maketitle
\thispagestyle{empty}
\pagestyle{empty}

\begin{abstract}
The efficient integration of multisensory observations is a key property of the brain that yields the robust interaction with the environment.
However, artificial multisensory perception remains an open issue especially in situations of sensory uncertainty and conflicts.
In this work, we extend previous studies on audio-visual (AV) conflict resolution in complex environments.
In particular, we focus on quantitatively assessing the contribution of semantic congruency during an AV spatial localization task.
In addition to conflicts in the spatial domain (i.e. spatially misaligned stimuli), we consider gender-specific conflicts with male and female avatars.
Our results suggest that while semantically related stimuli affect the magnitude of the visual bias (perceptually shifting the location of the sound towards a semantically congruent visual cue), humans still strongly rely on environmental statistics to solve AV conflicts.
Together with previously reported results, this work contributes to a better understanding of how multisensory integration and conflict resolution can be modelled in artificial agents and robots operating in real-world environments.
\end{abstract}
\section{INTRODUCTION}

Multisensory perception is crucial for a robust and efficient interaction with the environment~\cite{Stein93}.
The mammalian brain comprises a rich set of neural mechanisms that integrate multisensory stimuli on the basis of their physical properties (spatial and temporal alignment~\cite{Macaluso2004}) in combination with internal knowledge and expectations (e.g. semantic congruency~\cite{Laurienti2004}).
Noisy environmental conditions can lead to sensory uncertainty and decreased reliability, and thus require the interplay of both external and internal factors to efficiently trigger behaviourally relevant actions also in situations of a multisensory conflict~\cite{Polley2017}.

Robots operating in real-world environments can highly benefit from using robust mechanisms of multisensory perception with the goal to yield swift and singular behaviour also in situations of sensory uncertainty and conflict~\cite{CruzIROS}.
However, while computational models of multisensory integration (mostly audio-visual) have been extensively studied in the literature by considering either exogenous factors (spatial and temporal alignment) and endogenous factors (e.g. object identity) separately, the intricate interplay of such factors has been poorly understood.
In fact, the majority of the proposed approaches learn multisensory representations from congruent (co-occurring) audio-visual pairs (e.g.,~\cite{Parisi2016}\cite{Torralba2017}\cite{Zisserman2017}), whereas they do not account for handling situations of conflict (i.e. the visual and the auditory channel convey discordant information).

Behavioural studies have shown multiple audio-visual effects that reflect the interplay of spatial and semantic information processing (e.g., the spatial ventriloquism effect~\cite{Jack73}, where the location of the auditory stimulus is perceptually shifted towards the position of the synchronous visual one).
However, the oversimplified stimuli used by these studies (light blobs and beeps) do not reflect the complexity of the multisensory stimuli that robots are typically exposed to in natural environments.
Therefore, interdisciplinary efforts are required in order to provide insights into how artificial agents and robots should behave in situations of multisensory conflict in complex environments.

To better model multisensory conflict resolution in robots, we have designed and conducted a series of behavioural studies consisting of a localization task from audio-visual complex stimuli~\cite{ParisiICDL},\cite{ParisiIROS}.
In contradistinction to previously proposed studies using oversimplified stimuli, our studies were conducted in an immersive projection environment and comprised a scene with four animated avatars which can produce spatially (in)congruent lip movements, body motion, and vocalizations.
We observed that environmental statistics embedded in the scene modulate the magnitude and extension (in terms of the integration window) of the visually-induced bias.
In other words, when an avatar is moving its lips, the perception of a spatially incongruent vocalized sound is shifted towards the position of this avatar~\cite{ParisiICDL}.
Furthermore, we found that the integration of statistically relevant cues (moving lips and vocalized sounds) are quite robust to perceptually salient stimuli which act as a distractor (an avatar moving its arm but not the lips)~\cite{ParisiIROS}.

In this work, we extend the previously proposed studies by introducing an additional component to the localization task: semantic congruency.
While in the previous studies all of the four avatars had the same appearance, in this novel study we consider three male-looking avatars and one female-looking one.
The vocalizations can also be either male or female.
In this way, we aim at quantitatively assessing the contribution of semantic congruency to multisensory interaction and conflict resolution in complex environments.

In the following sections, we introduce the behavioural study, analyze the collected data, and discuss the obtained results from the perspective of multisensory modelling for artificial agents and robots.

\section{BEHAVIOURAL STUDY}

\subsection{Participants}

A total of 32 subjects (aged 21--35, right-handed, 6 female) participated in our experiment.
All participants reported that they had normal or corrected-to-normal vision and hearing and did not have a history of neurological conditions such as seizures, epilepsy, and stroke.
All participants signed a consent form approved by the Ethics Committee of Universit\"at Hamburg.
This study was conducted in accordance to the principles expressed in the Declaration of Helsinki.

\subsection{Apparatus, Stimuli, and Procedure}

\begin{figure}[t]
\centering
\includegraphics[width=0.47\textwidth]{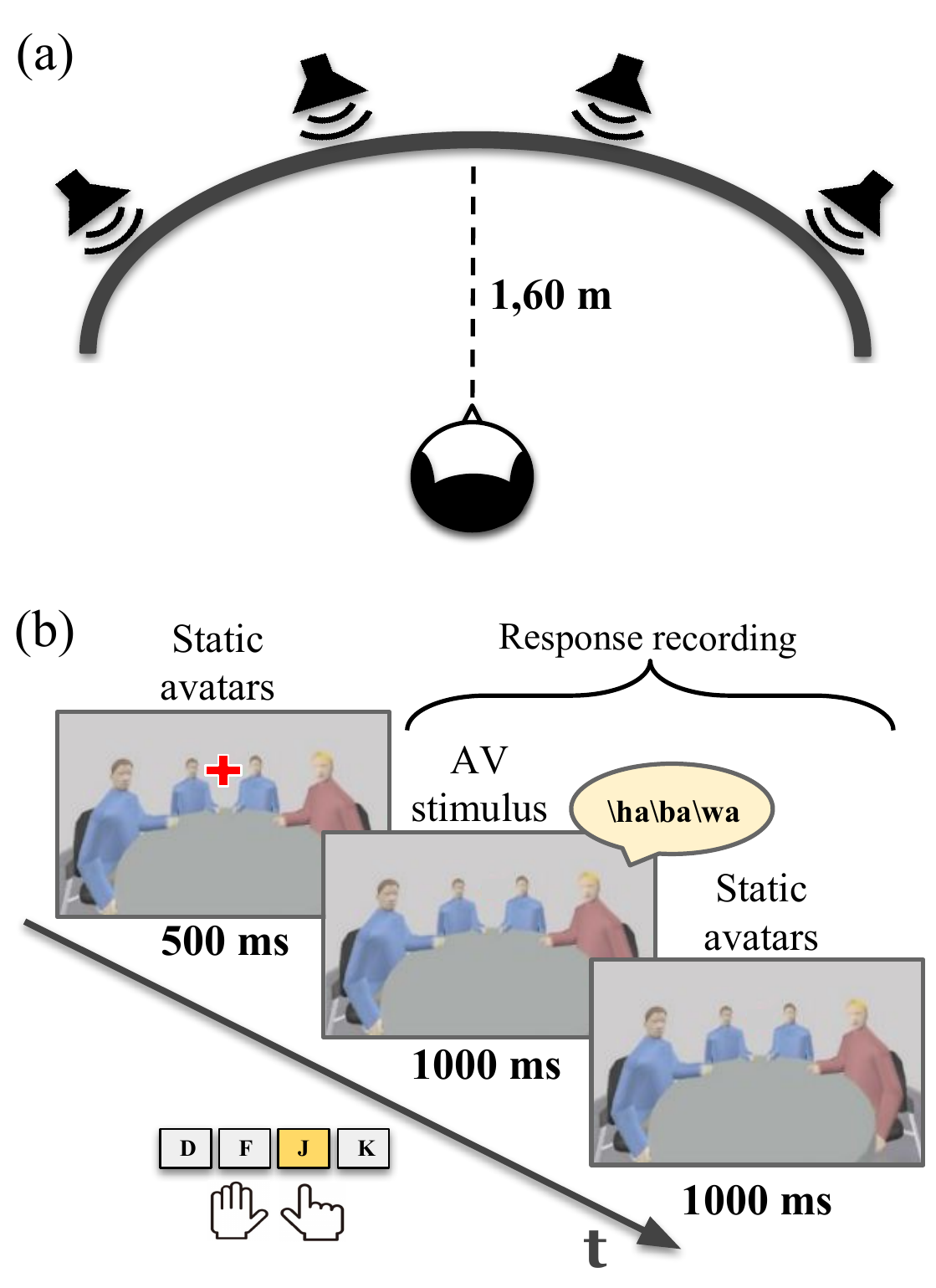}
\caption{Behavioural study on audio-visual localization: (a) Immersive experimental set-up with concave projection screen (b) Schematic illustration of one trial of the AV localization task.}
\label{fig:experiment}
\end{figure}

The participants sat at a desk with their chins positioned on a chin-rest 160 cm from the projection screen.
Visual stimuli were presented on a concave projection with four free-field speakers behind it that differed in azimuth and were placed to match the mouths of the projected avatars~(Fig~\ref{fig:experiment}.a).
The mouths of the avatars were 1.36 m above the ground and the avatars' locations were at -33, -11, +11, and +33 degrees off the centre (fixation point).
For a more detailed description of the experimental set-up, see our previous work~\cite{ParisiIROS}.

The AV localization task consisted of the participants having to select at each trial which avatar they believe the auditory stimulus was coming from.
One of the four avatars would move his/her lips along with a spatially congruent or incongruent vocalization.
We refer to the avatar moving its lips as the \textit{animated} avatar and \textit{static} otherwise.
The auditory stimulus consists of the combination of 3 syllables (all permutations without repetition composed of "ha", "wa", "ba") vocalized by a male and a female human voice.
Each trial comprises a female-looking avatar (FA) and three male-looking avatars (MA) with one animated avatar~(Fig~\ref{fig:experiment}.b).
The spatial location of the FA is pseudo-randomized across trials.
The duration of both the visual and auditory stimuli is 1000 ms.

We analyzed the following types of conflict:
\begin{enumerate}
\item \textit{Voice-Lips Spatial Conflict}: The location of the animated avatar (either FA or MA) is either congruent or incongruent to the location of the vocalized sound (either male or female voice).
\item \textit{Voice-Lips Gender Conflict}:  Gender-specific incongruency, i.e., animated FA with male voice or animated MA with female voice.
\item \textit{Voice-Appearance Gender Conflict}: Static or animated avatar (FA or MA) with a gender-incongruent, spatially-congruent voice.
\end{enumerate}

To ensure that participants have understood the instructions, they began the experiment with 12 practice trials composed of congruent AV stimuli.
The formal task consisted of 768 trials distributed across 3 sessions.
A schematic illustration of one trial is shown in~Fig~\ref{fig:experiment}.b.
Each trial started with static avatars and a fixation point displayed for 500 ms, followed by an AV stimulus and then another 1000 ms with static avatars.
The subjects were asked to produce a response within 2000 ms after the onset of the AV stimulus.

\subsection{Results and Analysis}

We analyzed the data in terms of the error rate (ER) of each participant with respect to the ground-truth position of the auditory stimulus.
Since visual spatial resolution is higher in the center of the field of view (FOV), the magnitude of the visual bias is expected to be stronger towards the center rather than towards the periphery~\cite{Odegaard15}.
Therefore, we also analyzed the results taking into account the distance between the avatars and their absolute location with respect to the FOV of the observer.
In line with our previous studies, divided the conditions in terms of (mirrored) spatial relationships:
\begin{itemize}
\item \textit{Central}: The two avatars in the center.
\item \textit{Lateral}: The two avatars on the right or the left side.
\item \textit{1-avatar gap}: Two avatars having a 1-avatar gap.
\item \textit{2-avatar gap}: The two avatars at each side of the screen.
\end{itemize}

\begin{figure}[t]
\centering
\includegraphics[width=0.47\textwidth]{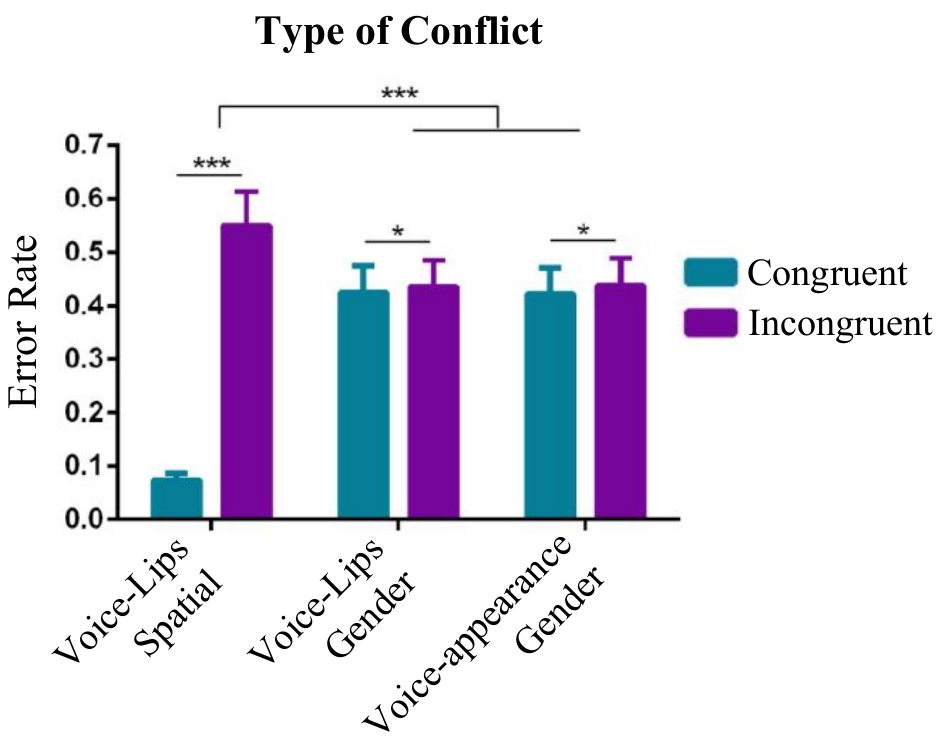}
\caption{Error rate (with respect to the ground-truth location of the auditory stimulus) for the different types of conflict. (* denotes $0.01 < p < 0.05$ and *** $p < 0.001$, error bars denote the standard deviation.)}
\label{fig:experiment1}
\end{figure}

\begin{figure*}[h]
\centering
\includegraphics[width=0.95\textwidth]{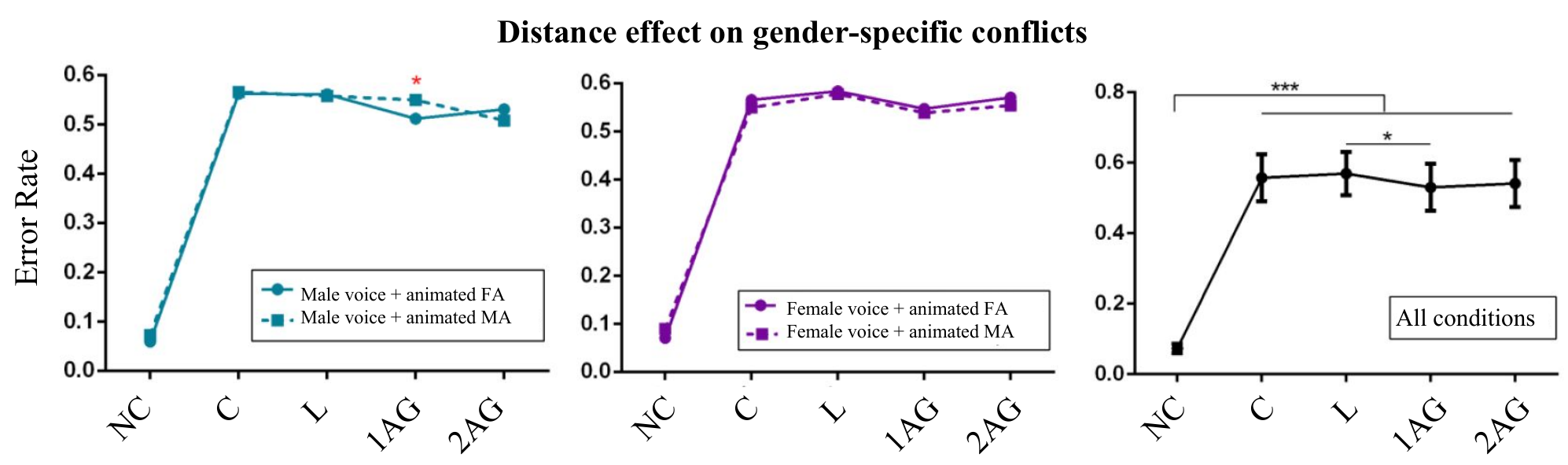}
\caption{ER analysis of the distance effect on gender-specific conflicts: NC (No conflict), C (central), L (lateral), 1AG (1-avatar gap), 2AG (2-avatar gap). (* denotes $0.01 < p < 0.05$ and *** $p < 0.001$, error bars denote the standard deviation.)}
\label{fig:experiment2}
\end{figure*}

For the three types of conflict, we conducted a repeated measures analysis of variance (ANOVA; shown in Fig.~\ref{fig:experiment1}).
The main effect of the different types of conflict was significant ($F=57.48$, $p<0.001$, $\eta^2=0.65$) and the main effect of the congruency was significant ($F=66.41$, $p<0.001$, $\eta^2=0.68$).
Post hoc \textit{t}-tests showed that the ER for the \textit{voice-lips spatial conflict} ($ER = 31.2\%$) was significantly lower than the \textit{voice-lips gender conflict} ($ER =43.1\%$, $p < 0.001$) and \textit{voice-appearance gender conflict} ($ER = 43.1\%$, $p < 0.001$).
Furthermore, there was no significant difference between these latter two conditions ($p > 0.05$).
For the congruency main effect, the participants' responses were less accurate under incongruent conditions ($ER = 47.5\%$) compared to congruent ones ($ER = 30.7\%$, $p < 0.001$).
There was a significant interaction effect between the type of conflict and congruency ($F = 52.04$, $p < 0.001$, $\eta^2 = 0.63$), showing a significantly larger effect in the \textit{voice-lips spatial conflict}.

The main effect of \textit{gender-specific voice conflict} ($F = 5.46$, $p < 0.05$, $\eta^2 = 0.15$) and the distance effect ($F = 50.49$, $p < 0.001$, $\eta^2 = 0.62$) were significant~(see Fig.~\ref{fig:experiment2}).
The ER for the female voice (ER=46\%) was significantly higher than for the male voice ($ER = 45\%$).
Regarding the distance effect, the congruent condition yielded the highest accuracy ($ER=7\%$), which is significantly higher than the other conditions ($p < 0.001$).
Post hoc \textit{t}-tests showed that the ER for the \textit{1-avatar gap} ($ER=54\%$) was lower than for the \textit{laterals} ($ER = 57\%$, $p <0.05$).
There was no significant interaction effect.
We conducted the repeated-measures ANOVA again by separating the analysis into two different voice genders (male voice with animated MA/FA and female voice with animated MA/FA).
We found that, when the voice is male, the interaction effect between the gender of the animated avatar and the distance was marginally significant ($F = 2.33$, $p = 0.08$, $\eta^2 = 0.07$).
The simple effect analysis showed that, when the participants heard a male voice, even though the animated FA was 1-avatar apart from the ground-truth sound location, humans' errors were still significantly higher ($ER = 55\%$) than when observing an animated MA ($ER = 51\%$, $p < 0.05$).

\begin{figure*}[h]
\centering
\includegraphics[width=0.85\textwidth]{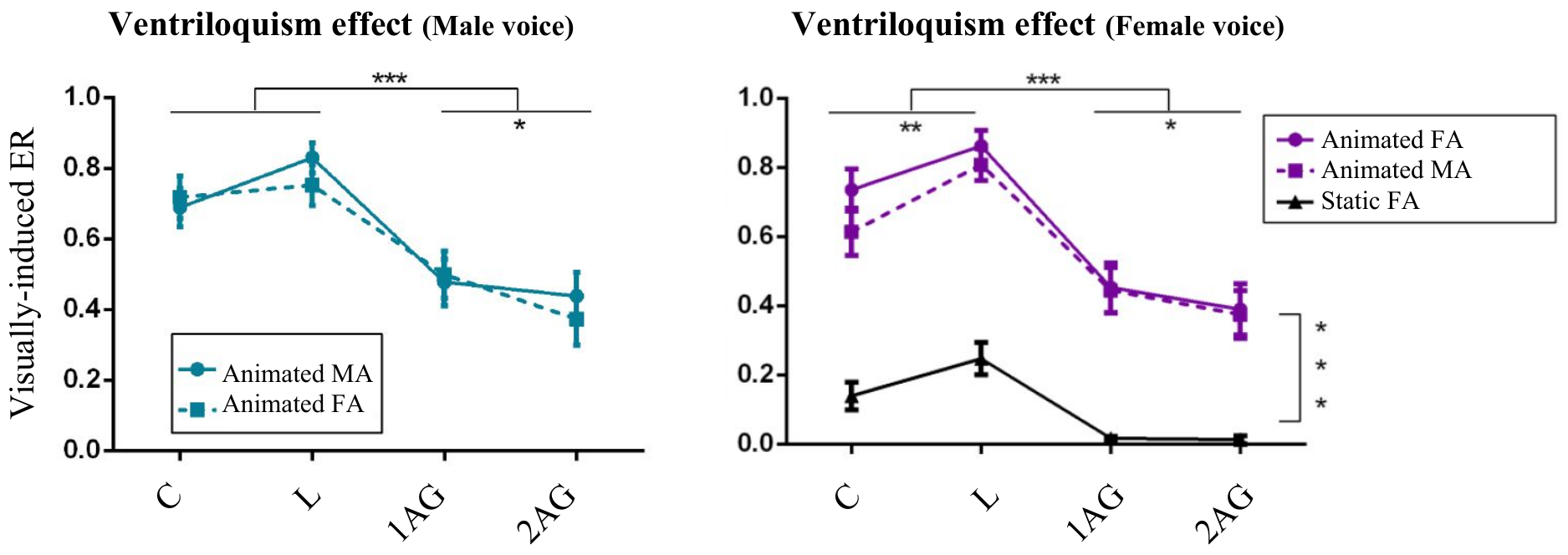}
\caption{Magnitude of the ventriloquism effect (visually-induced ER) for male and female voices: NC (No conflict), C (central), L (lateral), 1AG (1-avatar gap), 2AG (2-avatar gap). (* denotes $0.01 < p < 0.05$ and *** $p < 0.001$, error bars denote the standard deviation.)}
\label{fig:ventriloquist}
\end{figure*}

Additionally, we analyzed how the presence of a FA (either static or animated) biases the responses when the female voice is spatially congruent with a MA.
The main effect of the voice's gender was significant ($F=7.82$, $p < 0.01$, $\eta^2 = 0.20$).
We conducted repeated-measures ANOVA again by separating the analysis into two different voice gender (MA with male voice and MA with female voice).
In the analysis of the MA with male voice, neither the main effect of the visual cues and distance were significant, however, the interaction effect between these two factors was significant ($F = 4.99$, $p < 0.01$, $\eta^2 = 0.14$).
The simple effect analysis showed that for \textit{1-avatar gap}, the static FA (ER =56\%) had significantly higher ER than the animated FA (ER=52\%, $p < 0.01$).
For \textit{2-avatar gap}, static FA (ER=57\%) had significantly higher ER than the animated FA (ER = 51\%, $p < 0.05$).
In the analysis of MA with female voice, both the main effect of the visual cues and distance were not significant, with the interaction effect between these two factors being significant ($F = 3.82$, $p < 0.01$, $\eta^2 = 0.11$).

\subsection*{Ventriloquism Effect}

In our previous studies, we found that different types of visual stimuli have a different influence on the magnitude of the ventriloquism effect, i.e., the perceptual shift of the location of the auditory stimulus towards a visual bias~\cite{ParisiICDL},\cite{ParisiIROS}.
Here, we analyzed the ventriloquism effect under two independent conditions: male and female voice~(see Fig.~\ref{fig:ventriloquist}).

For the \textit{male voice} condition, the main effect of the distance was significant ($F = 28.96$, $p < 0.001$, $\eta^2 = 0.48$).
Post hoc \textit{t}-tests showed that the ER for \textit{central} ($ER = 70\%$) and \textit{lateral} ($ER = 79\%$) were significantly higher than for the \textit{1-avatar gap} ($ER = 49\%$, $p <0.001$) and \textit{2-avatar gap} ($ER=41\%$, $p <0.001$).
For the \textit{female voice} condition, the main effect of the visual bias ($F = 53.51$, $p < 0.001$, $\eta^2 = 0.64$) and distance ($F = 33.95$, $p < 0.001$, $\eta^2 = 0.53$) were significant.
Post hoc \textit{t}-tests showed that the magnitude of the bias from a static FA ($ER = 10\%$) was significantly weaker than from an animated FA ($ER = 61\%$) and animated MA ($ER = 57\%$, $p <0.001$).
ER under the \textit{central} ($ER = 50\%$) and \textit{lateral} ($ER = 64\%$) conditions had significantly higher ER than the \textit{1-avatar gap} ($ER = 32\%$, $p <0.001$) and \textit{2-avatar gap} ($ER = 27\%$, $p <0.001$).
There was a significant interaction effect between these two factors ($F = 5.61$, $p < 0.01$, $\eta^2 = 0.16$).
The simple effect analysis showed that under each distance condition, the visual bias driven by a static FA was significantly weaker than the one by animated avatars ($p <0.001$).
These results suggest that although semantically related (gender-specific) cues affect the magnitude of the visual bias, humans still strongly rely on environmental statistics to solve AV conflicts, i.e., tending to perceptually bind a vocalized sound to an animated avatar.


\section{CONCLUSION AND FUTURE WORK}

In this work, we extended previously proposed studies to quantitatively assess the contribution of semantic congruency during an AV spatial localization task.
We considered gender-specific congruency with female and male avatars and its relation with spatial conflicts (misaligned AV pairs).
Our obtained results suggest that humans still strongly rely on environmental statistics to solve AV conflicts while semantically-related stimuli affect the magnitude of the visual bias to a lesser extent.

Future studies should consider stronger semantic information, e.g. person identity and unique facial features.
While gender-specific conflicts influence the magnitude of the visual bias, one could argue that stronger semantic relations could have a stronger influence and, speculatively, dominate over other factors.

From a modelling perspective, we showed in a similar study that a humanoid robot can reproduce human-like behaviour when exposed to similar experimental conditions than humans through the use of a deep neural network architecture~\cite{ParisiIROS}.
However, the architecture was trained in an supervised fashion (i.e. using the human behaviour as the desired output), whereas it would be interesting for robots to learn the robust behaviour via unsupervised learning.

\addtolength{\textheight}{-12cm}   



%

\section*{ACKNOWLEDGMENT}

\small{This research was supported by National Natural Science Foundation of China (NSFC), the China Scholarship Council, and the German Research Foundation (DFG) under project Transregio Crossmodal Learning (TRR 169).}


\balance
\bibliographystyle{IEEEtran}
\bibliography{cmlbiblio18}

\end{document}